%% file: acl.tex
\pdfoutput=1

\documentclass[11pt]{article}

\usepackage[]{EMNLP2022}

\usepackage{times}
\usepackage{latexsym}

\usepackage[T1]{fontenc}

\usepackage[utf8]{inputenc}

\usepackage[T1]{fontenc}    
\usepackage{hyperref}       
\usepackage{url}            
\usepackage{booktabs}       
\usepackage{amsfonts}       
\usepackage{amsmath}
\usepackage{nicefrac}       
\usepackage{microtype}      
\usepackage{xcolor}         
\usepackage{enumitem}

\usepackage{microtype}
\usepackage{booktabs}
\usepackage{graphicx}
\usepackage{xspace}
\usepackage{amsfonts}
\usepackage{amsmath}
\usepackage{multirow}
\usepackage[noend,ruled]{algorithm2e}

\newcommand\our{FFF-NER\xspace}

\def\@fnsymbol#1{\ensuremath{\ifcase#1\or *\or \dagger\or \ddagger\or
   \mathsection\or \mathparagraph\or \|\or **\or \dagger\dagger
   \or \ddagger\ddagger \else\@ctrerr\fi}}
\newcommand{\ssymbol}[1]{^{\@fnsymbol{#1}}}

\title{Formulating Few-shot Fine-tuning Towards Language Model Pre-training: \\ A Pilot Study on Named Entity Recognition}

\author{Zihan Wang $\ \ $ Kewen Zhao $\ \ $ Zilong Wang $\ \ $ Jingbo Shang\thanks{$\ \ $Corresponding Author.}\\
  University of California, San Diego \\
  \{ziw224, k4zhao, zlwang, jshang\}@ucsd.edu
}

\begin{document}
\maketitle

\begin{abstract}
\input{0-abstract}

\end{abstract}

\input{1-introduction}
\input{2-preliminary}
\input{3-method}
\input{4-experiments}
\input{6-conclusion}

\bibliography{anthology,custom}
\bibliographystyle{acl_natbib}

\clearpage
\appendix

\input{7-appendix}

\end{document}

%% file: 0-abstract.tex
Fine-tuning pre-trained language models is a common practice in building NLP models for various tasks, including the case with less supervision. 
We argue that under the few-shot setting, formulating fine-tuning closer to the pre-training objective shall be able to unleash more benefits from the pre-trained language models.
In this work, we take few-shot named entity recognition (NER) for a pilot study, where existing fine-tuning strategies are much different from pre-training.
We propose a novel \underline{f}ew-shot \underline{f}ine-tuning \underline{f}ramework for NER, \our.
Specifically, we introduce three new types of tokens, 
``\emph{is-entity}'', ``\emph{which-type}'' and \emph{bracket}, so we can formulate the NER fine-tuning as (masked) token prediction or generation, depending on the choice of the pre-training objective. 
In our experiments, we apply \our to fine-tune both BERT and BART for few-shot NER on several benchmark datasets and observe significant improvements over existing fine-tuning strategies, including sequence labeling, prototype meta-learning, and prompt-based approaches.
We further perform a series of ablation studies, showing few-shot NER performance is strongly correlated with the similarity between fine-tuning and pre-training.

%% file: 1-introduction.tex
\section{Introduction}
\begin{figure*}[t]
\vspace{-3mm}
    \centering
    \includegraphics[width=0.99\linewidth]{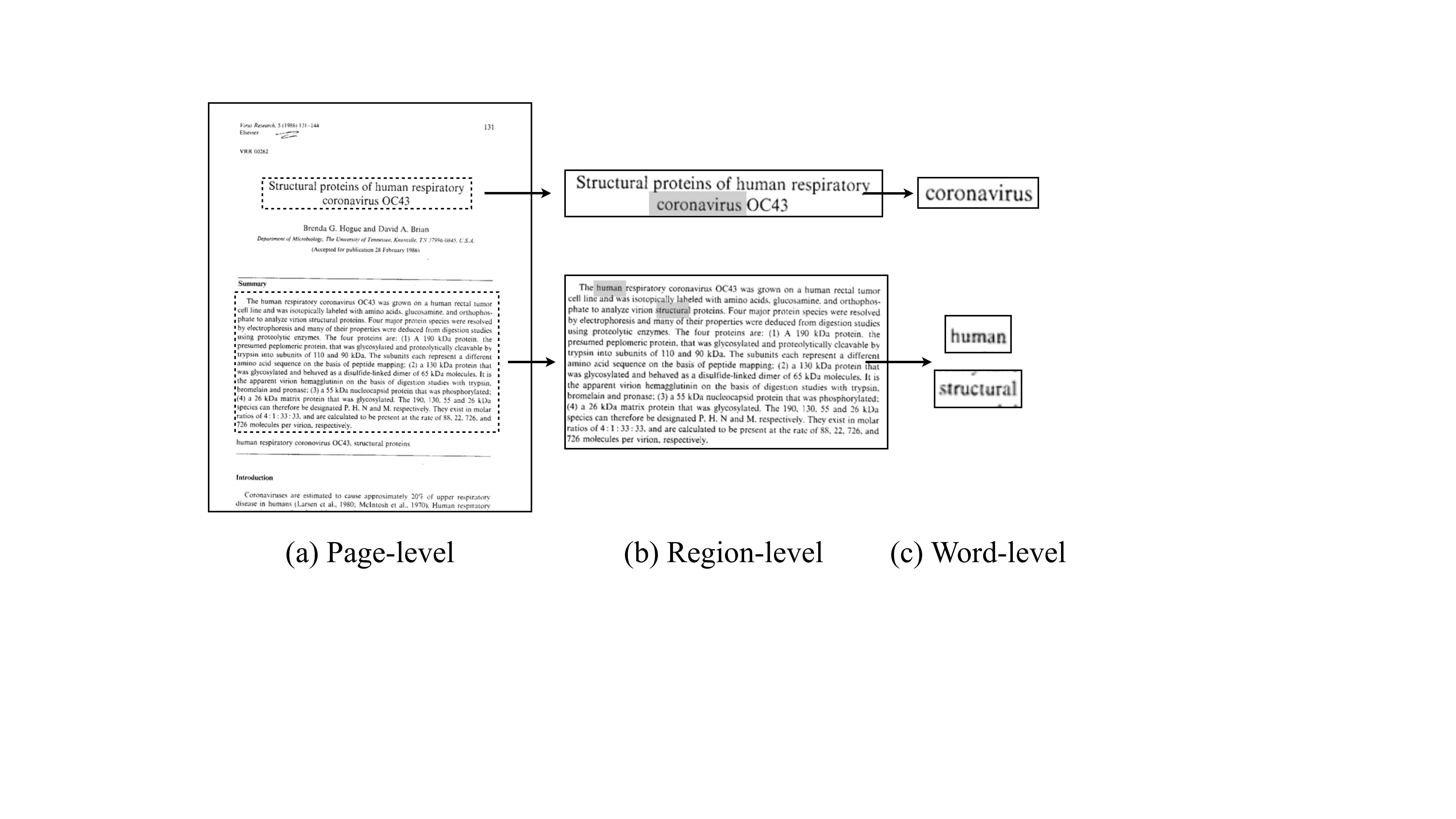}
    \caption{Two pre-training strategies (masked language modeling and sequence to sequence denoising) and different few-shot fine-tuning frameworks based on them. Our \our has different designs for different pre-trained models, but share the same principle of simulating the pre-training task. 
    }
    \label{fig:pipeline}
    \vspace{-3mm}
\end{figure*}

Pre-trained language models~\cite{DBLP:conf/naacl/PetersNIGCLZ18,DBLP:conf/naacl/DevlinCLT19,DBLP:journals/corr/abs-1907-11692,DBLP:conf/acl/LewisLGGMLSZ20} have the ability to capture extensive semantic and syntactic information in text. 
Such prior understanding of language offers a solid foundation to downstream tasks, 
therefore, fine-tuning language models have been widely applied in various NLP applications,
especially when the task-specific annotations are limited (a.k.a. few-shot setting)~\cite{radford2019language, DBLP:conf/nips/BrownMRSKDNSSAA20}.

While many few-shot strategies have been explored, there is no established systematic principle on designing these strategies.
Intuitively, given the same amount of task-specific annotations, a pre-training aligned fine-tuning strategy should more easily extract beneficial knowledge encoded during pre-training.
Therefore, we hypothesize that the performance of few-shot fine-tuning is tied to the similarity between the fine-tuning strategy and the pre-training objective. 

To verify this hypothesis, we take the Named Entity Recognition (NER) task for a pilot study, since the common fine-tuning strategies for NER are much different from popular pre-training tasks. 
Specifically, the most common fine-tuning paradigm for NER is to train a sequence labeling model on top of a pre-trained language model --- for each token in a sentence, simultaneously decide whether it belongs to an entity and the type of the entity among given entity types.
As an example showing the difference, the commonly used pre-trained language model BERT has a pre-training goal to disambiguate some masked tokens by predicting the correct token (masked language modeling). 
While in sequence labeling, \textit{each} token (representation) is to decide \textit{two} different pieces of information (span \& type). 

Following our hypothesis, we propose a few-shot fine-tuning framework for NER, \our as shown in Figure~\ref{fig:pipeline}. 
At the core is decoupling span detection and type prediction, since they are essentially two quite different tasks, which is not easy to learn altogether in few-shot fine-tuning. 
We introduce two new types of tokens that we dub as ``\emph{is-entity}'' and ``\emph{which-type}'' to perform the two tasks individually. 
With the help of these two types of tokens, we consider each instance not a single sentence but a sentence-span pair and ask the model to predict whether the span is an entity and, if it is, the type of the entity. 
The two tokens are placed around the span. 
By introducing another type of token, \emph{bracket}, around the span and the two special tokens, the sentence highlights the span in consideration and treats the special tokens as meta-data-like information. 
The goal of the fine-tuning task is then to classify the corresponding ``\emph{is-entity}'' and ``\emph{which-type}'' representation for each span.

Our \our is formulated in a pre-training agnostic way, and here we elaborate how it can be adapted to two popular pre-training tasks, (masked) token prediction and generation. 
For BERT trained with masked language modeling, we use the masked tokens as the two special tokens and ask the model to perform a binary classification on ``\emph{is-entity}'' and a multi-class classification on ``\emph{which-type}''. 
For BART trained with sequence to sequence denoising that is able to generate text, the two tokens are virtual --- we treat the formulated sentence with \emph{brackets} indicating spans and type-decoded special tokens as the target clean sentence, and the original sentence as a noised one without entity information. 
The model is to decode the target sentence with entity information. 

We conduct experiments on the recently proposed few-shot NER benchmark~\cite{DBLP:conf/emnlp/HuangLSJBCPG021} and show that \our can achieve significant improvements over the state-of-the-art methods, including sequence labeling, prototype meta-learning models, and prompt-based approaches.
It is worth mentioning that several recent NER methods~\cite{DBLP:conf/acl/CuiWLYZ21,DBLP:conf/emnlp/YangK20,DBLP:conf/emnlp/WangCZG21,DBLP:journals/corr/abs-2109-13532} have different experimental settings and dataset/data-split selections (even though the dataset names may be the same).
Therefore, for a fair comparison, we have carefully re-evaluated them under the same setting. 

Our extensive ablation studies have further verified the correlation between the performance of NER and the similarity between fine-tuning and pre-training.
Focusing on the masked language modeling pre-training task, the key efforts that we made in \our to close the gap between pre-training and fine-tuning (i.e., the usage of \emph{mask}, \emph{brackets}, ``\emph{is-entity}'', and ``\emph{which-type}'' tokens) are all crucial to the NER performance. 

All these empirical results are in favor of our hypothesis, shedding insights on design principles for fine-tuning strategies. 
Remarkably, there are existing models GENRE~\cite{DBLP:conf/iclr/CaoI0P21} and TANL~\cite{DBLP:conf/iclr/PaoliniAKMAASXS21} sharing the same design principle as \our by formulating BART fine-tuning as a token generation task.
They can be viewed as siblings of \our, serving as another strong evidence in favor of our hypothesis. 
Our contributions are summarized as follows.
\begin{itemize}[leftmargin=*,nosep]
    \item We hypothesize for better performance, (few-shot) fine-tuning should be designed similarly to the pre-training as much as possible. We further conduct an ablation study to verify this insight.
    \item Our pilot study on few-shot NER shows favorable results for our hypothesis. Specifically, we propose a fine-tuning framework for few-shot NER \our, which outperforms sequence labeling, protocol methods, and prompt-based learning models on the five benchmark datasets for few-shot NER.
    \item We enrich the standard few-shot NER benchmark by including more state-of-the-art methods.
\end{itemize}

\noindent\textbf{Reproducibility.} We will release our code on GitHub\footnote{\url{https://github.com/ZihanWangKi/fffner}}.

%% file: 2-preliminary.tex
\section{Related Works}

\subsection{Named Entity Recognition}
It has been shown that traditional fine-tuning methods that work in full supervision might not work well in few-shot learning scenarios~\cite{DBLP:conf/emnlp/WangCZG21}, as is the case in widely used BIO sequence labeling and near state of the art fully supervised NER models~\cite{DBLP:conf/acl/LiFMHWL20}. Some works have been proposed based on templates or prototypical networks targeting low resource NER~\cite{DBLP:conf/acl/CuiWLYZ21,DBLP:conf/emnlp/YangK20,DBLP:conf/emnlp/WangCZG21}. However, there lacks a general principle for few-shot NER. 
In this work, we claim that the similarity between the fine-tuning task and pre-training task can help few-shot learning and prove empirically with our pre-training tailored few-shot fine-tuning design and several ablations. 

\subsection{Prompt Learning}
Recently, a resource-less way of solving downstream tasks, prompt learning~\cite{DBLP:journals/corr/abs-2107-13586} has gained interest. By stirring knowledge from pre-trained language models with manually designed templates, the GPT model~\cite{radford2019language, DBLP:conf/nips/BrownMRSKDNSSAA20} can achieve decent performance on downstream tasks without training. For NER, since the labeling task is unseen in natural text, prompt-based methods are often combined with further fine-tuning~\cite{DBLP:conf/acl/CuiWLYZ21}. Our idea goes along the same line, where we argue that the similarity to the pre-training task can help the model better transfer knowledge from pre-training to the fine-tuning task. We show that some templates~\cite{DBLP:conf/acl/CuiWLYZ21} might not work well for sequence-to-sequence denoising trained models, as they are still much different from the pre-training task.

\section{Preliminaries}

\subsection{Problem Definition}
The task of Named Entity Recognition is to extract the n-grams in the text that are named entities and label them with a set of pre-defined entity types. 
Formally, given a sequence of words $w_1, w_2, \ldots, w_n$ of length $n$, the goal is to identify the spans $w_l, w_{l+1}, \ldots, w_r$ where $l \leq r$ that is an entity, and assign it an entity type $c$ from a set of pre-defined entity types $C$ (e.g., $C = \{\text{Person, Location, etc}\}$). In this work, we sometimes use $l\ldots r$ to denote the span for simplicity.

\paragraph{Metric.}
The most widely used metric for NER is the (span-based, micro) F$_1$ score. It considers all typed spans in the ground truth and from the model's prediction and calculates its F$_1$ score on retrieving the whole entities.

\paragraph{Few-shot.}
There are different definitions of Few-shot learning for NER. In this work, we focus on the N-way K-shot setting~\cite{DBLP:conf/emnlp/HuangLSJBCPG021}, where we consider all $N = |C|$ entity types and for each entity type, randomly pick $K$ sentences with supervision that contains the particular entity type. This is to make a fair comparison with the traditional sequence labeling methods since full supervision of sentences is crucial. Undoubtedly, while we select $N * K$ sentences as supervision, the total number of entity spans in the sentences can be greater than $N * K$. Therefore, it is important to evaluate all models on the same randomly selected sets of sentences. And one major reason for us choosing this setting is the established benchmark over ten NER datasets~\cite{DBLP:conf/emnlp/HuangLSJBCPG021}.

\subsection{Pre-trained Language Model}
The use of a pre-trained language model has become ubiquitous in NLP. By training with self-supervision, often through disambiguating corrupted sentences, pre-trained language models can obtain a general understanding of the text. Studies have shown that the pre-training task is more a major factor for downstream task performance, rather than the architecture~\cite{DBLP:journals/corr/abs-2205-05131}. 

\noindent\textbf{BERT}~\cite{DBLP:conf/naacl/DevlinCLT19} is a transformer~\cite{DBLP:conf/nips/VaswaniSPUJGKP17} encoder model trained with masked language modeling. After replacing some tokens with a special mask token, the model attempts to recover the original tokens. Thus, BERT can leverage the nearby context of the mask token to predict its origin. 

\noindent\textbf{BART}~\cite{DBLP:conf/acl/LewisLGGMLSZ20} is a transformer encoder-decoder model trained with sequence to sequence denoising. The whole sentence is corrupted by replacing segments with masks, and the model is to recover the original uncorrupted sentence. A BART can predict the complete, clean sentence given a partial sentence.

%% file: 3-method.tex
\section{Our Framework \our}
In this section, we introduce our framework \our, and specifically talk about how to adapt it for a masked language modeling and sequence to sequence pre-trained model. 

\subsection{Task Formation}
The goal is to design a fine-tuning task equivalent to NER and similar to pre-training, where the goal is to disambiguate sentences.
We consider each possible span in a sentence and predict whether it is an entity and, if it is, the type of it. The major difference from our task to a traditional sequence labeling task is that we decouple the span detection and entity prediction problems.

We introduce two tokens, ``is-entity'' and ``which-type'' for span detection and type prediction. For a particular sentence and a span specified, we insert the two tokens around the span and then brackets around the two tokens and the span. As an illustration, suppose the sentence in consideration is 
{
\[ \small \text{{Tom lives in Los Angeles}} \]
}
and the span is ``Los Angeles'', then our task translates (by inserting 6 brackets and 2 special tokens) the sentence into 
{
\[ \small \text{{Tom lives in [ is-entity ] [ Los Angeles ] [ which-type ]}} \]
}
The brackets ``highlight'' the two special tokens and the span in the sentence, which are important for understanding whether the span is an entity and the type of it. The brackets may also hint that the special tokens should be treated as meta-data, instead of normal text. 

Different pre-trained language models will have slightly different training objectives and prediction strategies based on such a task formation. We will focus on the two popular pre-trained language models, BERT and BART, and talk about integrating them with the framework.

\subsection{Instantiation of FFF-NER using BERT}
We first talk about BERT, an encoder model pre-trained with masked language modeling.

\paragraph{Model Architecture.}
We use the BERT model to encode the translated sentence. Then, the last layer representations for the two special tokens are extracted, and each is fed through a classification head to retrieve the class logits for both span detection and type prediction. The pre-trained model is also tuned (i.e., the parameters are not frozen), as in many fine-tuning approaches. We use the mask token for both ``is-entity'' and ``which-type''.

\paragraph{Training. }
We train the model to detect whether the span is an entity and predict the entity type for each span sentence pair. We consider a negative sampling approach to treat the given entities as positive instances and all the other spans as potential negative ones. Specifically, for each sentence $W = w_1, w_2, \ldots, w_n$ with typed entity spans $\mathcal{E}_W = \{ (l_1, r_1, t_1), (l_2, r_2, t_2), \ldots, (s_e, r_e, t_e) \}$ that are given as supervision, we consider a ``positive'' loss for the sentence and a certain entity span
\begin{equation*}
    \mathcal{L}^\text{pos} = \mathcal{L}(\text{is-entity}) + \mathcal{L}(\text{which-type}),
\end{equation*}
where $\mathcal{L}(\text{is-entity})$ is the cross-entropy loss for the binary prediction problem of classifying the ``is-entity'' token, and $\mathcal{L}(\text{which-type})$ is the cross-entropy loss for the multi-class prediction problem of classifying the exact type. The ``negative'' loss is learned on spans that are not entities. For the sentence $W$, there are $|W| + \binom{|W|}{2} - |\mathcal{E}_W|$ spans that are not entities. We consider sampling a subset $\mathcal{D}_W$ of negative spans for each sentence every epoch. The sample can change every epoch, which gives high coverage over all the negative spans. Then, for each negative span, the loss is 
\begin{equation*}
    \mathcal{L}^\text{neg} = \mathcal{L}(\text{is-entity}).
\end{equation*}
Here, we only consider the is-entity loss since there is no entity type for the sampled negative span.

During training in each epoch, we first sample all the negative samples along with the positive ones, treat it as the dataset to train on in the epoch, and perform regular (mini-)batch gradient descent on the dataset with the loss.
\begin{equation*}
    \mathcal{L} = \mathcal{L}^\text{pos} + \mathcal{L}^\text{neg}.
\end{equation*}
\paragraph{Negative Sampling for Training. }
Here, we detail how we sample the negatives during training. There are two things to consider, the number of negative samples to pick and the probability of each span being sampled as a negative.

Directly training on all negative samples is very hard, since there are much more negatives than positives, resulting in an imbalanced training set. We describe how we determine the number of negatives to pick: the number of negative samples depends on the length of the sentence and the number of entities given as supervision in it. Intuitively, the longer the sentence is, the more negative spans we need to pick to cover every token that appears in it. And the more entities in the sentence, the more negative spans we need to pick to cover overlapping non-entities with the entities. We consider these two measurements separately. 

For sentence length, we consider $\alpha * |W|$ negative samples where $\alpha$ is a small constant. This is based on a statistical result where $O(|W| \log |W|)$ randomly sampled spans can cover each word with high probability, and an empirical reasoning that since sentences are short, even after removing the log factor, the sampled spans still can cover almost all words in expectation.

We then take a simple approach where we \textit{project} entities as virtual words that increases the length of the sequence; then, we fall back to the formula proposed for the sentence lengths. Based on statistics in the CoNLL~\cite{DBLP:conf/conll/SangM03} training set, the ratio of tokens to the number of entities is about $10: 1$. We count each entity as $10$ words and select $\alpha * 10 * |\mathcal{E}_W|$ negative samples.

Finally, we sum up these two numbers as the negative samples for each sentence. We empirically show that this choice works well, and through sensitivity studies, the choice of $\alpha$ is not significant to performance changes.

We also note that since we only sample a portion of negatives, we should consider a weighted sampling since there are important negatives: those that have high overlap with a true entity (e.g., “Chicago is” and “York”). The model learns spurious features without explicitly treating them as negatives.

To solve this, for each sentence, we define a sampling probability based on the overlap between the non-positive spans and entities. If the span contains words $w_l, w_{l+1}, \ldots, w_r$, and among these $r - l + 1$ words, $c$ of them are within an entity in the sentence, the probability for the span being sampled is proportional to $exp(\frac{c}{r - l + 1})$. The probability of each sample is determined based on the intuition that the hard instances to predict are those that look like an entity. Therefore, we calculate the overlap between the span and the entities, normalized by the length of the span to assign a priority score for each span. Simply treating this score (after normalizing to a distribution) as the probabilities to sample will cause the sampling process never to pick spans that have no intersection with entities since their priority score is 0. So, we transform the percentage into an exponential scale and normalize thereafter to form a probability distribution.

\paragraph{Predicting.}
During prediction, we enumerate through all spans in the sentence and extract the probability of it being classified as an entity $p_{l, r}$ (from the ``is-entity'' token), along with the type (the class that has the maximum probability on \textit{which-type} token). Ideally, all the spans with $p_{l, r} \geq 0.5$ should be disjoint since all our positive instance training is performed on non-overlapping spans from a non-nested NER dataset. However, training is hard to be perfect, especially in few-shot scenarios. Therefore, we need a strategy to resolve the overlapping predicted spans. 

We consider the most straightforward way to resolve overlaps: greedily assign spans based on their probability and ignore overlapping ones with lower probability. Formally, we consider all spans $(l, r)$ with $p_{l, r} \geq 0.5$ and sort them with the largest probability one first. Then, we pick the spans one by one and add them to the final predicted entity list, given that the span does not overlap with an already picked span, in which case we ignore it and move on to the next span.

\subsection{Instantiation of FFF-NER using BART}
MLM pre-training is not the only type of model that can be tuned under our setting. We talk about instantiating our framework with an encoder-decoder model pre-trained with sequence to sequence denoising, BART.


\paragraph{Setup}
The sequence denoising task is to recover the original sentence given a partially corrupted version. As a direct simulation of the MLM-based setting for FFF-NER, we can consider the original sentence as partially corrupted, and ask the model to generate the full sentence that contains the decoded ``is-entity'' and ``which-type'' tokens indicating the span and type. However, in sequence decoding, the ``is-entity'' part is unnecessary since the brackets are generated to predict the span. We illustrate the sentence denoising setup with an example.
Consider an input sentence 
{
\[ \small \text{Tom lives in Los Angeles} \]
}
then, a correctly decoded sentence is
{
\[ \small \text{[ Tom ] [ Person ] lives in [ Los Angeles ] [ Location ]} \]
}

We introduce two similar designs in literature.
\noindent\textbf{GENRE. }
GENRE~\cite{DBLP:conf/iclr/CaoI0P21} was originally proposed for entity linking. The setup is almost equivalent to our setup above
except for the use of different type of brackets.

\noindent\textbf{TANL. }
TANL~\cite{DBLP:conf/iclr/PaoliniAKMAASXS21} has a slightly different design that joints the adjacent brackets between the entity and the type with a ``|''. For the same input sentence example in the setup section, the correctly decoded sentence is 
{
\[ \small \text{[ Tom | Person ] lives in [ Los Angeles | Location ]} \]
}
Because of the existence of such work, our experiments will focus on the MLM-based models and we leave discussion of such sequence-to-sequence models in the Appendix.

%% file: 4-experiments.tex
\section{Experiments}

\input{tables/full-5shot}
\input{tables/main-5shot}

\subsection{Compared Methods}
We compare with several publicly available low-resource NER models. 

\noindent\textbf{Roberta-base} is a sequence labeling model build on the official \texttt{roberta-base}~\cite{DBLP:journals/corr/abs-1907-11692}. It is a baseline result reported in \citet{DBLP:conf/emnlp/HuangLSJBCPG021}. We also use BERT models and experiment with \textbf{BERT-\{base, large\}-\{uncased, cased\}}.

\noindent\textbf{Nearest Neighbor} is a baseline prototype method reported in \citet{DBLP:conf/emnlp/HuangLSJBCPG021} that assigns each token representation to the nearest label representation, learned by the examples in the training set in a prototypical network~\cite{DBLP:conf/nips/SnellSZ17}.

\noindent\textbf{StructShot}~\cite{DBLP:conf/emnlp/YangK20} extends Nearest Neighbor to consider label transitions as in a Conditional Random Field~\cite{DBLP:conf/icml/LaffertyMP01}. 

\noindent\textbf{TemplateNER}~\cite{DBLP:conf/acl/CuiWLYZ21} considers templates that are human selected formatted like ``X is a Y entity .'' and ``X is not a named entity .'' where X is the span to consider and Y is the ground truth entity during training. It uses BART to decode the template for each sentence span pair. 

\noindent\textbf{EntLM}~\cite{DBLP:journals/corr/abs-2109-13532} also has the idea of minimizing the gap between pre-training and fine-tuning. They asked a MLM-pretrained language model to predict every token in the input sentence. The prediction should either be within list of special tokens specifying each type if it is an entity or the original token itself otherwise. This setting draws the pre-training and fine-tuning closer by making token predictions, but poses another difference in that the target predicted word might not suit the context well. In fact, they need to specially curate a list of entity names. More discussions on this are in the Appendix.

\noindent\textbf{SpanNER}~\cite{DBLP:conf/emnlp/WangCZG21} is a method that also breaks span detection and type prediction into two tasks. For each token, they simultaneously predict whether it is a start of span and an end, similar to question answering. Then, they utilize class descriptions from annotation guidelines or Wikipedia to construct a class representation to match the detected spans. In some sense, their ideology follows our framework, as an important step in our framework is also decoupling span detection and type prediction. However, many of their model designs still differ from the backbone pre-trained model BERT (e.g., asking each token to perform span predictions and averaging span representations to find the closest class representation).

\subsection{Experimental Settings}
For the four BERT sequence labeling models, we use the implementation of sequence labeling in Huggingface~\cite{DBLP:conf/emnlp/WolfDSCDMCRLFDS20,huggingface}, and tune the number of epochs to use. 

We reproduce the other compared methods as their performance is not reported on the standard few-shot setting/standard datasets or splits in \citet{DBLP:conf/emnlp/HuangLSJBCPG021}. When reproducing, we follow the original hyper-parameters except (1) Increasing the sequence length when necessary, (2) Decreasing the batch size if the GPU memory does not fit, and (3) Tuning the number of training epochs and learning rate if performance is not desirable.

Our masked language modeling based \our is built with Pytorch Lightning~\cite{pytorch_lightning}. We try to follow all hyper-parameter settings (optimizer, learning rate, etc.) as close as Huggingface's sequence labeling. 

All experiments are conducted on RTX A6000 and RTX 8000 GPUs, both having 48 GB GPU RAM. 
For exact hyperparameter settings, please refer to the Appendix.

\subsection{Datasets}
We mainly experiment on five of the ten datasets in \citet{DBLP:conf/emnlp/HuangLSJBCPG021}. The reasons of choosing these five datasets are: (1) SpanNER~\cite{DBLP:conf/emnlp/WangCZG21} require descriptions that were not gathered for the other five datasets (2) These five datasets are commonly used and cover various domains (news, general, social media, reviews). We do include a study of our framework on all of the ten datasets. For each dataset, we run on the standard ten splits from \citet{DBLP:conf/emnlp/HuangLSJBCPG021} whenever possible and report both the average and standard deviation.

\subsection{Performance of \our}
In Table~\ref{tab:full-5shot} we show the performance of our framework on a BERT-based-uncased model on all ten datasets in \citet{DBLP:conf/emnlp/HuangLSJBCPG021} and compare with the two baselines that do not require extra data: sequence labeling and prototype. We see that our model performs statistically significantly better than the compared methods on every dataset except WNUT17. We also note that our overall performance on the ten datasets exceeds the best method reported in \citet{DBLP:conf/emnlp/HuangLSJBCPG021} (LC + NSP + ST) that has an average performance of 58.5. Notably, we do not require a large-scale noisy fine-grained entity recognition dataset to adapt the pre-trained language model or self-training on the unlabeled in-domain dataset. 

In Table~\ref{tab:main-5shot} we show the full table containing our method and all the compared methods on the five datasets performing 5-shot NER, differentiating by the underlying pre-trained language model. 
Each baseline performed pretty well on some of the datasets, especially EntLM and SpanNER, which we attribute to their unanimous design of bridging pre-training and fine-tuning. However, none of them showed consistent improvement across all datasets like ours. When looking at the absolute scores, our large methods are statistically significantly better than the compared methods on all datasets except Onto.

\subsection{Ablations of \our}
\input{tables/ablations}
We strengthen our claim that the improvement of our method comes from the fine-tuning design being similar to the pre-training task. We consider three variants of our framework design for masked language modeling and show the results in Table~\ref{tab:ablations}.

\noindent\textbf{Not mask} replaces the mask token used for the two special tokens with the actual tokens ``span'' and ``type''. We can see that the performance drops significantly on CoNLL, while no significant changes can be observed from Restaurant. We attribute this to the fact that there is less difference from a mask token to an actual token for pre-trained language models since MLM is also trained on non-mask tokens for $20\%$ of the training time~\cite{DBLP:conf/naacl/DevlinCLT19}. However, the slight decrease in performance still indicates that using the mask token is better.

\noindent\textbf{No brackets} removes all six brackets when formulating the input sentence. Since we have the two mask tokens around the span, the input sentence is still distinct for different spans. However, the performance drops by a large margin. At first, this may sound intriguing since the sentence without brackets looks more like natural text. However, we note that with the original setting with brackets, the sentence is like annotating a span with meta-data, while without brackets, the model is to complete the sentence with missing words.

\noindent\textbf{Span \& type together} removes the ``is-entity'' token and integrates the span detection and type prediction task. The ``which-type'' token now needs to perform two tasks simultaneously, much like in sequence labeling. The performance on the two datasets drops drastically: on CoNLL, the performance is similar to that of sequence labeling, and on MIT Restaurant, the performance is even lower.

\subsection{A Mismatched Scenario of \our}
\input{tables/mismatched.tex}
Given our hypothesis that pre-training fine-tuning gap should be minimized, one would expect that a framework designed to reduce the gap between MLM pre-training shouldn't work well for Sequence Denoising pre-training. We show that this is indeed the case with an experiment in Table~\ref{tab:mismatched}. Even though \our outperforms Sequence Labeling on an MLM model, as expected, it does not show a difference on a model with a different pre-training objective.
This observation can be even strengthened if we can check the other way -- whether Sequence Denoising tailored methods fall short on MLM pre-trained models. However, we note that all such methods we tested in this work require the ability to generate text, which MLM models can not.

\subsection{Sensitivity Study of \our}
\input{tables/sensitivity}
We study the choice of the number of negative spans to sample in Figure~\ref{fig:sensitivity}. In our framework, we sample $\alpha * (|W| + |\mathcal{E}_W| * 10)$ negative spans for each sentence, and $\alpha$ is chosen to be 3 across all datasets. While we fix the use of $10$, we analyze how the performance varies with different choices of $\alpha$. We can see that overall, the performance is pretty steady even when we use two times smaller number of samples ($\alpha = 1$) or almost one time more number of samples $(\alpha = 5)$. Therefore, we believe that one does not need extensive hyper-parameter selection on $\alpha$.



\subsection{Performance vs Shots}
\input{tables/moreshots}
We also study how our framework works with more supervision. In Figure~\ref{fig:shots} we increase the number of shots $K$ from the default $5$ shots to as many as $100$ shots. A clear trend is that the advantages of our framework vanish as the number of shots increases. We also train \our with \texttt{BERT-base-uncased} on the full CoNLL dataset, achieving a performance of 95.57 on the dev set and 92.06 on the test set. This result is slightly less but close to a sequence labeling based BERT model.  
The vanishing performance in high-resource scenarios aligns with our common sense --- the gap between pre-training and fine-tuning can be overcome when there is large amounts of supervision. Overall, our framework is still capable of handling large amounts of supervision, albeit designed for a few-shot scenario.

%% file: tables/full-5shot.tex
\begin{table}[t]
    \centering
    \small
    {
    \begin{tabular}{l c c c}
    \toprule
    \textbf{Dataset} & \textbf{LC} & \textbf{P} & \textbf{\our}  \\
    \midrule
    CoNLL & 53.5 & 58.4 & \textbf{67.90} ($\pm 3.95$) \\
    \midrule
    Onto & 57.7 & 53.3 & \textbf{66.40} ($\pm 1.62$) \\
    \midrule
    WikiGold & 47.0 & 51.1 & \textbf{61.42} ($\pm 8.82$)\\
    \midrule
    WNUT17 & 25.7 & 29.5 & 30.48 ($\pm 3.69$) \\
    \midrule
    MIT Movie & 51.3 & 38.0 & \textbf{60.32} ($\pm 1.12$) \\
    \midrule
    MIT Restaurant & 48.7 & 44.1 & \textbf{51.99} ($\pm 2.33$) \\
    \midrule
    SNIPS & 79.2 & 75.0 & \textbf{83.43} ($\pm 1.44$)\\
    \midrule
    ATIS & 90.8 & 84.2 & \textbf{92.75} ($\pm 0.99$)\\
    \midrule
    Multiwoz & 12.3 & 21.9 & \textbf{47.56} ($\pm 8.46$)\\
    \midrule
    I2B2 & 36.0 & 32.0 & \textbf{49.37} ($\pm 5.08$)\\
    \midrule
    \midrule
    Average & 50.2 & 48.8 & \textbf{61.15}\\
    \bottomrule
    \end{tabular}
    }
    \caption{Experiments on N-way 5 shot NER on all datasets in \citet{DBLP:conf/emnlp/HuangLSJBCPG021} for our model. LC stands for a sequence labeling method, P stands for prototypical network, and \our is a BERT-base-uncased model trained with our framework. We bold the best scores that are statistically significant when compared with others at a 0.05 significance level.}
    \label{tab:full-5shot}
    \vspace{-6mm}
\end{table}


%% file: tables/main-5shot.tex
\begin{table*}[t]
    \centering
    \small
    \resizebox{\textwidth}{!}{
    \begin{tabular}{l l c c c c c}
    \toprule
    \textbf{Framework} & \textbf{Language Model} & \textbf{CoNLL} & \textbf{Onto} & \textbf{WNUT17} & \textbf{MIT Movie} & \textbf{MIT Restaurant}  \\
    \midrule
    \multirow{8}{*}{Seq Labeling} & Roberta-base$\ssymbol{2}$ & 53.5 & 57.7 & 25.7 & 51.3 & 48.7 \\
    \cmidrule(lr){2-7}
    & BERT-base-uncased & 52.92 ($\pm 4.55$) & 56.66 ($\pm 0.85$) & 21.53 ($\pm 4.81$) & 47.29 ($\pm 1.32$) & 45.64 ($\pm 2.30$) \\
    \cmidrule(lr){2-7}
    & BERT-base-cased & 51.04 ($\pm 3.94$) & 60.71 ($\pm 1.11$) & 20.18 ($\pm 5.43$) & 44.98 ($\pm 1.48$) & 39.07 ($\pm 2.73$) \\
    \cmidrule(lr){2-7}
    & BERT-large-uncased & 53.50 ($\pm 4.40$) & 59.65 ($\pm 1.87$) & 28.03 ($\pm 5.08$) & 52.84 ($\pm 2.00$) & 47.33 ($\pm 1.91$) \\
    \cmidrule(lr){2-7}
    & BERT-large-cased & 52.23 ($\pm 3.90$) & 64.03 ($\pm 1.28$) & 26.76 ($\pm 4.30$) & 48.31 ($\pm 2.93$) & 43.08 ($\pm 1.98$)\\
    \midrule
    \multirow{1}{*}{Prototype} & Roberta-base$\ssymbol{2}$ & 58.4 & 53.3 & 29.5 & 38.0 & 44.1 \\
    \midrule
    TemplateNER & BART-large & 61.46 ($\pm 6.46$) & N/A & 11.20 ($\pm 1.09$) & 41.80 ($\pm 1.36$) & N/A \\
    \midrule
    StructShot & BERT-large-cased & 50.55 ($\pm 7.75$) & 69.80 ($\pm 1.19$) & 27.32 ($\pm 3.01$) & 56.58 ($\pm 2.01$) & 45.62 ($\pm 3.39$)\\
    \midrule
    \multirow{3}{*}{EntLM$\ssymbol{3}$} & BERT-base-cased & 61.67 ($\pm 2.72$) & 67.97 ($\pm 1.24$) & 23.70 ($\pm 4.03$) & 43.95 ($\pm 0.79$) & 51.29 ($\pm 2.12$) \\
    \cmidrule(lr){2-7}
    & BERT-large-cased & 59.44 ($\pm 2.85$) & 68.92 ($\pm 1.58$) & 17.82 ($\pm 3.06$) & 45.31 ($\pm 6.09$) & 53.13 ($\pm 1.56$) \\
    \midrule
    \multirow{3}{*}{SpanNER$\ssymbol{3}$} & BERT-base-uncased & 58.95 ($\pm 4.15$) & 67.89 ($\pm 1.33$) & 22.18 ($\pm 4.63$) & 55.79 ($\pm 1.22$) & 52.08 ($\pm 2.46$) \\
    \cmidrule(lr){2-7}
    & BERT-large-uncased & 57.93 ($\pm 3.99$) & 68.40 ($\pm 1.39$) & 20.72 ($\pm 6.58$) & 57.81 ($\pm 2.29$) & 50.10 ($\pm 1.06$) \\
    \midrule
    \multirow{3}{*}{\our} & BERT-base-uncased & 67.90 ($\pm 3.95$) & 66.40 ($\pm 1.62$) & 30.48 ($\pm 3.69$) & 60.32 ($\pm 1.12$) & 51.99 ($\pm 2.33$) \\
    \cmidrule(lr){2-7}
    & BERT-large-uncased & 69.23 ($\pm 3.90$) & 69.43 ($\pm 1.30$) & 34.96 ($\pm 5.07$) & 61.31 ($\pm 2.12$) & 55.01 ($\pm 2.71$) \\
    \bottomrule
    \end{tabular}
    }
    \caption{Experiments on N-way 5 shot NER on 5 datasets. The average performance over ten different folds of seed shots are given, and the standard deviation is given in parentheses.  $\ssymbol{2}$ indicates results borrowed from \citet{DBLP:conf/emnlp/HuangLSJBCPG021}, and they did not report standard deviations. N/A indicates the performance is far behind other methods, and we suspect that extensive dataset-specific template/hyperparameter tuning is necessary. For the backbone language model, we show the better one among the cased and uncased versions. We discuss the reproduction details and number differences from reported numbers of EntLM and SpanNER in the Appendix.}
    \label{tab:main-5shot}
    \vspace{-3mm}
\end{table*}

%% file: tables/ablations.tex
\begin{table}[t]
    \centering
    \small
    \begin{tabular}{l c c}
    \toprule
    \textbf{Framework} & \textbf{CoNLL} & \textbf{MIT Restaurant} \\
    \midrule
    Sequence Labeling & 52.92 ($\pm 4.55$) & 45.64 ($\pm 2.30$) \\
    \our & 67.90 ($\pm 3.95$) & 51.99 ($\pm 2.33$) \\
    \ \ Not mask & 64.08 ($\pm 3.62$) & 52.71 ($\pm 3.47$) \\
    \ \ No brackets & 60.18 ($\pm 3.42$) & 42.73 ($\pm 4.21$) \\
    \ \ Span \& type together & 53.20 ($\pm 7.96$) & 33.79 ($\pm 6.87$) \\
    \bottomrule
    \end{tabular}
    \caption{Ablations of \our that distance the fine-tuning task from pre-training. We use a \texttt{BERT-base-uncased} model. }
    \label{tab:ablations}
    \vspace{-3mm}
\end{table}

%% file: tables/mismatched.tex
\begin{table}[t]
    \centering
    \small
    \begin{tabular}{l c c}
    \toprule
    \textbf{Backbone} & \textbf{BERT} & \textbf{BART} \\
    \midrule
    Sequence Labeling & 52.92 ($\pm 4.55$) & 59.66 ($\pm 2.92$) \\
    \our for MLM & 67.90 ($\pm 3.95$) & 59.65 ($\pm 2.61$) \\
    \bottomrule
    \end{tabular}
    \caption{An experiment where \our for Masked Language Modeling trained models (e.g., BERT) are applied to Sequence Denoising trained models (e.g., BART).}
    \label{tab:mismatched}
    \vspace{-3mm}
\end{table}

%% file: tables/sensitivity.tex
\begin{figure}
    \centering
    \includegraphics[width=0.99\linewidth]{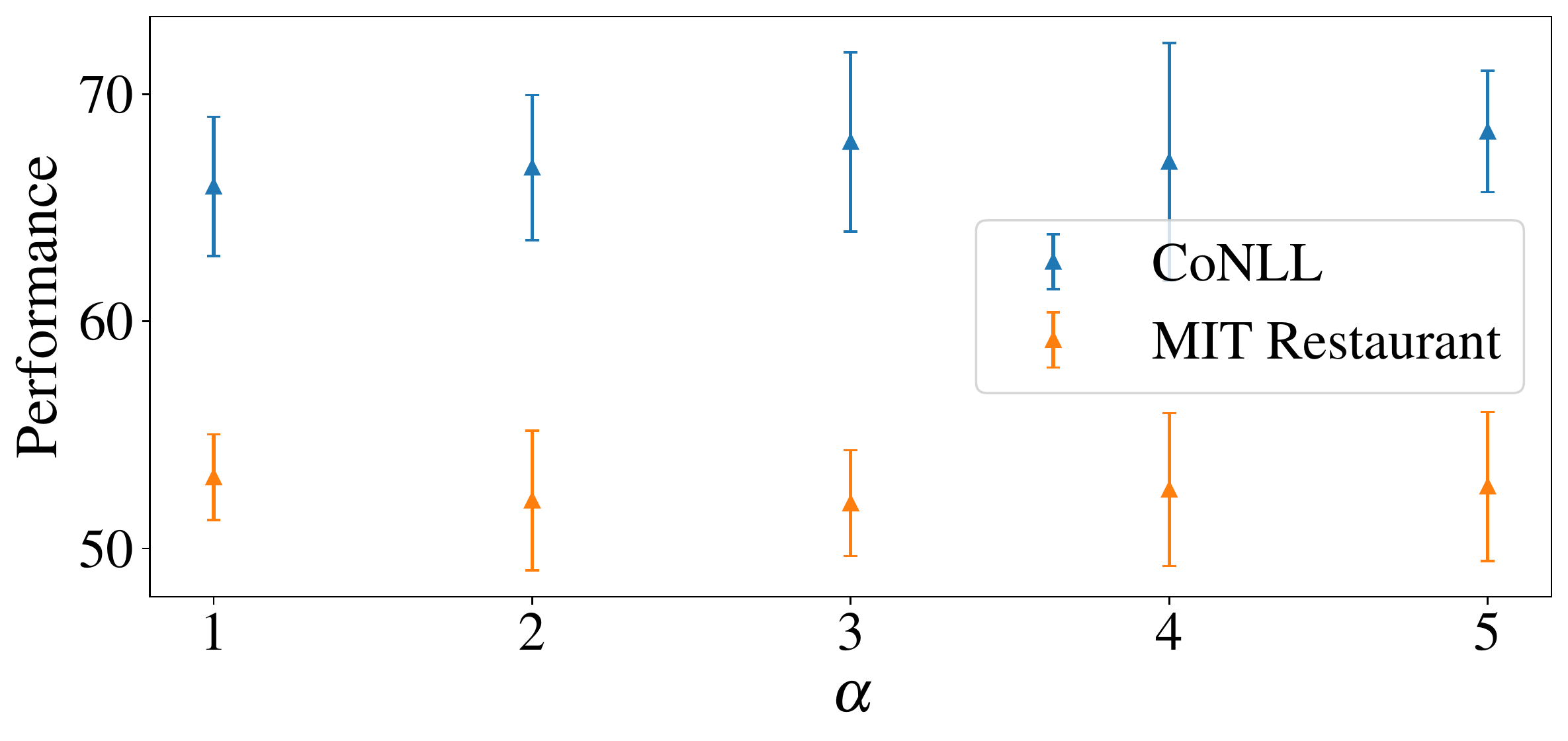}
    \caption{Sensitivity study using on CoNLL and MIT Restaurant using \texttt{BERT-base-uncased}. Average performance and standard deviation are shown. }
    \label{fig:sensitivity}
    \vspace{-3mm}
\end{figure}

%% file: tables/moreshots.tex
\begin{figure}
    \centering
    \includegraphics[width=0.99\linewidth]{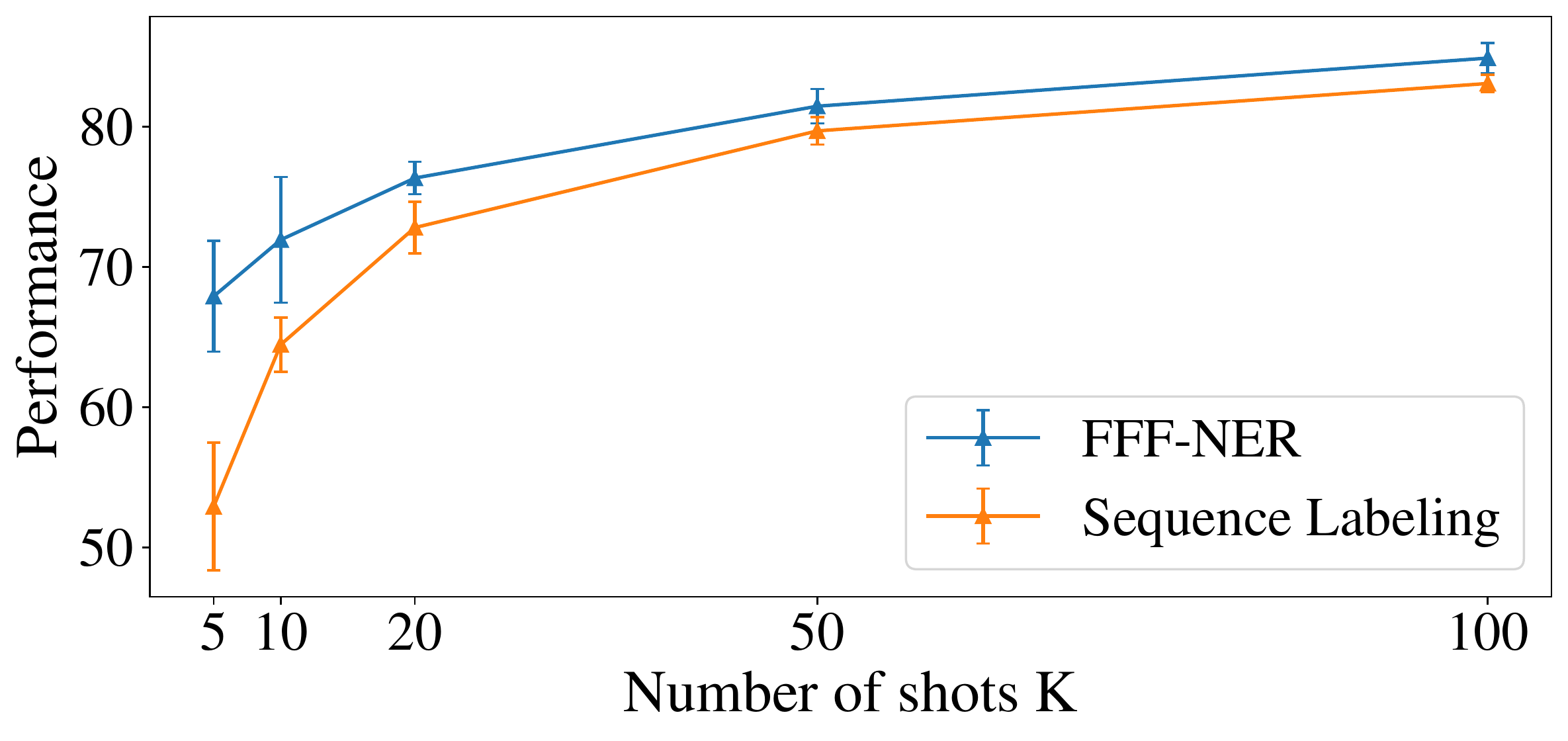}
    \caption{Analysis of our \our when the number of shots vary on the CoNLL dataset using \texttt{BERT-base-uncased}. Average performance and standard deviation are shown. A sequence labeling model is also shown for comparison.}
    \label{fig:shots}
    \vspace{-3mm}
\end{figure}

%% file: 6-conclusion.tex
\section{Conclusions}
In this work, we draw attention to an intuitive principle for few-shot fine-tuning designs: the closer the fine-tuning task is to the pre-training task, the better the performance. 
We empirically prove this by designing a few-shot fine-tuning framework for NER, \our. Our framework is not constrained to a specific architecture of language models and can be easily extended to distinct pre-trained models (e.g., BERT and BART in our experiments). 
Our model outperforms existing fine-tuning strategies including sequence labeling, prototype meta-learning, and prompt-based approaches on the standard benchmark for few-shot NER. Through a series of ablations, we also show that if we manually remove ingredients in our framework that make the fine-tuning task similar to pre-training, the performance drops. 

We see future works in two directions. One is to incorporate label semantics into the BERT based \our, so that even less supervision can suffice. The other is to verify our principle of few-shot fine-tuning on other downstream tasks.


\section*{Ethical Considerations}
The work presented in this paper deals with design principles in the few-shot scenario. We present experiments on Named Entity Recognition, which do not pose ethical concerns. Further, we will also open source our code. Our framework makes it possible for training NER models with small amounts of supervision, which makes NER tools more accessible to every ordinary people, so we are on the positive side on ethical consideration.

\section*{Limitations}
\paragraph{Time Complexity.}
One drawback of our \our design for MLM models is the increased time complexity. Compared to a sequence labeling method which needs one forward pass to predict entities of the whole sentence, our requires potentially $n + \binom{n}{2}$ passes. However, a few things to note is: 
\begin{itemize}[leftmargin=*,nosep]
    \item If one has prior knowledge on how long the entity could be, they can only consider spans with at most that length and avoid the quadratic dependency on the sequence length.
    \item In few-shot learning, the training set is small, so the increase in training time is insignificant.
\end{itemize}
Yet, even with these considerations, our work is less efficient than approaches that require a single pass for each sentence, like the traditional sequence labeling method.

\paragraph{Special Treatment for Large Training Sets. }
Shown in the Appendix, our method needs a smaller learning rate when we experiment with more data (e.g., 100 shot scenario).
Otherwise, there will be cases when the model training fails due to gradient vanishing 
We also observed this behavior for EntLM~\cite{DBLP:journals/corr/abs-2109-13532} and tuned the learning rate for it as well. We do not think that this problem is severe since (1) they are noticeable without a development set as one can check the gradients and the training accuracy/loss to detect such problems and (2) decreasing learning rate or restarting with different initializations can resolve this problem in our experiments.

\section*{Acknowledgement}
We thank all anonymous reviewers and program chairs for their helpful feedback. Zihan Wang is supported by the UCSD Jacob School of Engineering Fellowship and the UCSD Hal\i c\i o\u glu Data Science Fellowship. This work is sponsored in part by National Science Foundation Convergence Accelerator under award OIA-2040727 as well as generous gifts from Google, Adobe, and Teradata. Any opinions, findings, and conclusions or recommendations expressed herein are those of the authors and should not be interpreted as necessarily representing the views, either expressed or implied, of the U.S. Government. The U.S. Government is authorized to reproduce and distribute reprints for government purposes not withstanding any copyright annotation hereon.

%% file: 7-appendix.tex
\section{\our for MLM}
We include some pseudo code pieces to help understand the whole framework when applied to masked language modeling based pre-training language model.

\SetAlgoSkip{}
\begin{algorithm}[h]
    \caption{Create} \label{alg:data}
    \textbf{Input}: Sentence $W$ and entities $\mathcal{E_{W}}$.\\
    $\overline{\mathcal{E_W}} =$ all spans not in $\mathcal{E_{W}}$. \\
    \If {is predicting} {
        \textbf{Return} Formulated Input for $(W, e), \forall e \in \mathcal{E_{W}} \cup \overline{\mathcal{E_W}}$. \\
    }
    \Else {
        $S(\overline{\mathcal{E_W}}) =$ sampled $\overline{\mathcal{E_W}}$. \\
        \textbf{Return} Formulated Input for $(W, e), \forall e \in \mathcal{E_{W}} \cup S(\overline{\mathcal{E_W}})$. \\
    }
\end{algorithm}
\begin{algorithm}[h]
    \caption{Training} \label{alg:train}
    \textbf{Input}: Few-shot Training Dataset $D_{train}$, Model $\mathcal{M}$. \\
    \For {each epoch} {
        $\overline{D_{train}} = \emptyset$ \\
        \For {each ($W, \mathcal{E_{W}}$) in $D_{train}$} {
            $\overline{D_{train}} = \overline{D_{train}} \cup Create(W, \mathcal{E_{W}})$\\
        }
        Train $\mathcal{M}$ on $\overline{D_{train}}$.\\
    }
    \textbf{Return} $\mathcal{M}$
\end{algorithm}
\begin{algorithm}[h]
    \caption{Predicting} \label{alg:predict}
    \textbf{Input}: Evaluation Dataset $D_{eval}$, Model $\mathcal{M}$. \\
    Final predictions $\mathcal{P} = \emptyset$ \\
    \For {each ($W, \mathcal{E_{W}}$) in $D_{eval}$} {
        $preds = $ $\mathcal{M}(Create(W, \mathcal{E_{W}}))$. \\
        $\mathcal{P} = \mathcal{P} \cup$ Resolve predictions $preds$. \\
    }
    \textbf{Return} $\mathcal{P}$
\end{algorithm}

\section{Dataset Statistics and License}
We refer the readers to the benchmark paper that standardized the datasets~\cite{DBLP:conf/emnlp/HuangLSJBCPG021}. 

\section{Hyperparameters}\label{sec:hyp}
We show the hyper-parameters we use for training the sequence labeling model and \our based on BERT in Table~\ref{tab:hyperparameters} for both base and large models. We note that there are some exceptions on learning rates for our framework. For the base models when experimenting with $100$ shots, we set the learning rate to $0.00001$; for the fully supervised training, we set the learning rate to $0.000002$. For the large models on the Onto dataset, we set the learning rate to $0.00001$. 
\begin{table}[t]
    \small
    \centering
    \begin{tabular}{l|c c}\toprule
                  & Sequence Labeling & \our-BERT \\\midrule
    Epochs        & 100/300 (base/large) & 30 \\\midrule
    Optimizer     & \multicolumn{2}{c}{AdamW} \\\midrule
    Learning Rate & \multicolumn{2}{c}{0.00002}$\ssymbol{2}$ \\\midrule
    Scheduler     & \multicolumn{2}{c}{Linear, no warmup} \\\midrule
    Batch Size    & \multicolumn{2}{c}{32} \\\midrule
    Dropout       & \multicolumn{2}{c}{0.1} \\\midrule
    $\alpha$      & N/A & 3 \\\bottomrule
    \end{tabular}
    \caption{Hyper-parameters for sequence labeling and \our-BERT, for all experiments in the paper. $\ssymbol{2}$ there is some exceptions for the learning rate, explained in text in Section~\ref{sec:hyp}.}
    \label{tab:hyperparameters}
\end{table}

\section{EntLM}
\begin{table*}[t]
    \centering
    \small
    \begin{tabular}{l l l | c}
    \toprule
    \textbf{Result Source} & \textbf{Model}        & \textbf{Label Source}   & \textbf{CoNLL} \\
    \midrule
    reported      & RoBERTa-base & Knowledge Base & 68.6 \\
    \midrule
    reproduced    & RoBERTa-base & Knowledge Base & 68.40 ($\pm 3.00$) \\
    \midrule
    reproduced    & RoBERTa-base & Ground Truth & 65.20 ($\pm 3.03$) \\
    \midrule
    reproduced    & BERT-base-cased & Ground Truth & 61.67 ($\pm 2.72$) \\
    \bottomrule
    \end{tabular}
    \caption{Reported and Reproduced numbers of EntLM with different model and label sources.}
    \label{tab:entlm}
\end{table*}
EntLM needs to identify representative tokens for each class. In their paper, they follow BOND~\cite{DBLP:conf/kdd/LiangYJEWZZ20} that labels the training set based on distant supervision from external knowledge bases. However, BOND was not developed for all the datasets in the benchmark, so we take an alternative approach where we give EntLM \textit{ground truth} full training set to find representative tokens.

Arguably, this should be an upper bound for EntLM since we replaced a noisy source to a clean source. However, we note that there is a performance difference of our reproduced EntLM from their paper. We ruled out the differences between theirs and ours in Table~\ref{tab:entlm}. We can see that while there is a difference emerging from the backbone model choice to our surprise, there is a performance drop when the source of labels comes from ground truth instead of a noisy distant supervision-based approach. We show the representative words in Table~\ref{tab:entlm_tokens} and note that the actual tokens selected are rather similar for the two approaches. We argue that EntLM might suffer from instability issues where a few small changes on the representative tokens (in our case, even trying to improve it with the ground truth source) can hurt the performance quite hard.

\begin{table*}[t]
    \centering
    \small
    \begin{tabular}{l | p{3cm} | p{3cm} | p{3cm} | p{3cm}}
    \toprule
    Source & Person & Organization & Location & Miscellaneous \\
    \midrule
    Distant Supervision & "Michael", "John", "David", "Thomas", "Martin", "Paul" &  "Corp", "Inc", "Commission", "Union", "Bank", "Party" & "England", "Germany", "Australia", "France", "Russia", "Italy" &"Palestinians", "Russian", "Chinese", "Dutch", "Russians", "English" \\
    \midrule
    Ground Truth     & "John", "Michael", "David", "Paul", "Thomas", "Mark" & "Corp", "National", "Inc", "Commission", "St", "Co" & "Australia", "France", "Germany", "England", "Italy", "Russia" & "German", "Australian", "British", "American", "European", "Israeli" \\
    \bottomrule
    \end{tabular}
    \caption{Representative tokens for classes in EntLM}
    \label{tab:entlm_tokens}
\end{table*}


\section{SpanNER}
Our reproduction of SpanNER on CoNLL2003 and MitMovie with 5-shot supervision is quite different from their original numbers. There are a few reasons behind this:
\begin{itemize}[leftmargin=*,nosep]
\item The dataset, particularly the labeling schema, of MiTMovie is different, while they used the same dataset names.
\item The choice of 5-shot supervision is likely different due to randomness. We followed the exact same splits in \citet{DBLP:conf/emnlp/HuangLSJBCPG021}. 
\item We reported average and std performance across 10 different random 5-shot supervision, while SpanNER reported average and std performance across 5 different model random initializations. The high variance we reported indicates that different choice of splits can lead to drastically different performance.
\end{itemize}
We believe that our choice of the following \citet{DBLP:conf/emnlp/HuangLSJBCPG021} has two advantages. First, this setting (datasets and data split) is open-sourced and can be compared with future work more easily. Second, the averaging across random splits make the result more stable since it accounts for the larger portion of the variance.

\section{Sequence Denoising Pre-training}
\input{tables/bart-main-5-shot}
We show the performance for models based on sequence denoising pre-trained models in Table~\ref{tab:bart-main-5shot}. We use GENRE and TANL to illustrate how \our works with BART since they are similar by design.
We note that it is quite hard to compare across different pre-training models, but both GENRE and TANL have established a strong performance improvement against the TemplateNER baseline and the sequence labeling and prototype baselines in \citet{DBLP:conf/emnlp/HuangLSJBCPG021}.

%% file: tables/bart-main-5-shot.tex
\begin{table*}[t]
    \centering
    \small
    \resizebox{\textwidth}{!}{
    \begin{tabular}{l l c c c c c}
    \toprule
    \textbf{Framework} & \textbf{Language Model} & \textbf{CoNLL} & \textbf{Onto} & \textbf{WNUT17} & \textbf{MIT Movie} & \textbf{MIT Restaurant}  \\
    \midrule
    Seq Labeling & Roberta-base$\ssymbol{2}$ & 53.5 & 57.7 & 25.7 & 51.3 & 48.7 \\
    \midrule
    Prototype & Roberta-base$\ssymbol{2}$ & 58.4 & 53.3 & 29.5 & 38.0 & 44.1 \\
    \midrule
    \multirow{3}{*}{\our} & BERT-base-uncased & 67.90 ($\pm 3.95$) & 66.40 ($\pm 1.62$) & 30.48 ($\pm 3.69$) & 60.32 ($\pm 1.12$) & 51.99 ($\pm 2.33$) \\
    \cmidrule(lr){2-7}
    & BERT-large-uncased & 69.23 ($\pm 3.90$) & 69.43 ($\pm 1.30$) & 34.96 ($\pm 5.07$) & 61.31 ($\pm 2.12$) & 55.01 ($\pm 2.71$) \\
    \midrule
    \midrule
    TemplateNER & BART-large & 61.46 ($\pm 6.46$) & N/A & 11.20 ($\pm 1.09$) & 41.80 ($\pm 1.36$) & N/A \\
    \midrule
    \multirow{3}{*}{\parbox{1.5cm}{\our (GENRE)}} & BART-base & 58.86 ($\pm 1.78$) & 56.12 ($\pm 0.87$) & 26.98 ($\pm 1.64$) & 52.45 ($\pm 2.01$) & 43.34 ($\pm 1.41$)  \\
    \cmidrule(lr){2-7}
     & BART-large & 71.00 ($\pm 2.67$) & 64.40 ($\pm 2.88$) & 41.48 ($\pm 1.58$) & 53.25 ($\pm 1.99$) & 51.00 ($\pm 1.52$)  \\
    \midrule
    \multirow{4}{*}{\parbox{1.5cm}{\our (TANL)}} & BART-base & 59.05 ($\pm 1.65$) & 62.13 ($\pm 0.97$) & 34.24 ($\pm 3.13$) & 55.61 ($\pm 1.42$) & 56.76 ($\pm 2.47$)  \\
    \cmidrule(lr){2-7}
     & BART-large & 60.32 ($\pm 2.47$) & 63.17 ($\pm 1.26$) & 36.98 ($\pm 1.37$) & 58.23 ($\pm 1.37$) & 59.84 ($\pm 2.27$)  \\
     \cmidrule(lr){2-7}
     & T5-base & 62.33 ($\pm 2.76$) & 69.27 ($\pm 1.40$) & 33.80 ($\pm 2.91$) & 56.46 ($\pm 1.49$) & 58.85 ($\pm 1.33$)  \\
    \bottomrule
    \end{tabular}
    }
    \caption{Experiments on N-way 5 shot NER on 5 datasets. The average performance over ten different folds of seed shots are given, and the standard deviation is given in parentheses.}
    \label{tab:bart-main-5shot}
    \vspace{-3mm}
\end{table*}